\documentclass{article}
\usepackage{ijcai22}

\usepackage{times}
\usepackage{soul}
\usepackage{url}
\usepackage[hidelinks]{hyperref}
\usepackage[utf8]{inputenc}
\usepackage[small]{caption}
\usepackage{graphicx}
\usepackage{amsmath}
\usepackage{amsthm}
\usepackage{booktabs}
\usepackage{algorithm}
\usepackage{algorithmic}
\usepackage{booktabs}
\usepackage{multirow}
\usepackage{rotating}
\usepackage{makecell}
\usepackage{color}
\usepackage{subcaption}
\usepackage{newfloat}
\usepackage{listings}
\usepackage{fancyhdr}
\usepackage{amssymb}

\title{Long-term Spatio-Temporal Forecasting via Dynamic Multiple-Graph Attention}

\author{
Wei Shao$^{1\dag,*}$
\and
Zhiling Jin$^{2\dag}$\and
Shuo Wang$^{3*}$\and
Yufan Kang$^4$\and
Xiao Xiao$^2$\and
Hamid Menouar$^5$\and
Zhaofeng Zhang$^1$\and
Junshan Zhang$^6$\And
Flora Salim$^7$
\affiliations
$^1$School of Electrical, Computer and Energy Engineering, Arizona State University, Tempe, United States\\
$^2$School of Telecommunications Engineering, Xidian University, Xi'an, China\\
$^3$D-ITET, ETH Zurich, Zurich, Switzerland\\
$^4$School of Computing Technologies, RMIT University, Melbourne, Australia\\
$^5$Qatar Mobility Innovations Center, Qatar University, Doha, Qatar\\
$^6$ECE Department at University of California Davis, Davis, United States\\
$^7$School of Computer Science and Engineering, University of New South Wales, Sydney, Australia
\emails
\{\href{mailto:phdweishao@gmail.com}{phdweishao}, \href{mailto:shawnwang.tech@gmail.com}{shawnwang.tech}\}@gmail.com,
\href{mailto:jinzhiling@stu.xidian.edu.cn}{jinzhiling@stu.xidian.edu.cn},
\href{mailto:s3711739@student.rmit.edu.au}{s3711739@student.rmit.edu.au},
\href{mailto:xiaoxiao@xidian.edu.cn}{xiaoxiao@xidian.edu.cn},
\href{mailto:hamidm@qmic.com}{hamidm@qmic.com},
\href{mailto:zzhan199@asu.edu}{zzhan199@asu.edu},
\href{mailto:jazh@ucdavis.edu}{jazh@ucdavis.edu},
\href{mailto:flora.salim@unsw.edu.au}{flora.salim@unsw.edu.au}
}

\begin{document}

\maketitle

\begin{abstract}
      Many real-world ubiquitous applications, such as parking recommendations and air pollution monitoring, benefit significantly from accurate long-term spatio-temporal forecasting (LSTF). LSTF makes use of long-term dependency structure between the spatial and temporal domains, as well as the contextual information. Recent studies have revealed the potential of multi-graph neural networks (MGNNs) to improve prediction performance. However, existing MGNN methods do not work well when applied to LSTF due to several issues: the low level of generality, insufficient use of contextual information, and the imbalanced graph fusion approach. To address these issues, we construct new graph models to represent the contextual information of each node and exploit the long-term spatio-temporal data dependency structure. To aggregate the information across multiple graphs, we propose a new dynamic multi-graph fusion module to characterize the correlations of nodes within a graph and the nodes across graphs via the spatial attention and graph attention mechanisms. Furthermore, we introduce a trainable weight tensor to indicate the importance of each node in different graphs. Extensive experiments on two large-scale datasets demonstrate that our proposed approaches significantly improve the performance of existing graph neural network models in LSTF prediction tasks. The code is available at \underline{\url{https://github.com/swsamleo/MLSTGCN}}.
\end{abstract}
\let\thefootnote\relax\footnotetext{$^{\dag}$Equal contribution.
\\\indent \hspace{5pt} $^*$Corresponding authors.} 
\section{Introduction}

Recently, various spatio-temporal prediction tasks have been investigated, including traffic flow \cite{li2018diffusion,ijcai2020-326,yu_spatio-temporal_2018}, parking availability \cite{zhang2020semi}, and air pollution \cite{wang2020pm25,wen2019novel,LIU2021101079}. All the scenarios above benefit from an accurate forecast by leveraging historical data in the long run, namely, long-term spatio-temporal forecasting (LSTF). 

% Recently, various spatio-temporal prediction tasks have been explored, including social network evolving \cite{min2021stgsn}, traffic flow prediction \cite{li2018diffusion,ijcai2020-326,yu_spatio-temporal_2018}, parking availability prediction \cite{zhang2020semi}, and air pollution prediction \cite{wang2020pm25,wen2019novel,LIU2021101079}. All the above scenarios require an accurate forecast by leveraging historical data in the long run: namely long-term spatio-temporal forecasting (LSTF). 

\begin{figure}[t]
      \centering
      \includegraphics[width=0.47\textwidth]{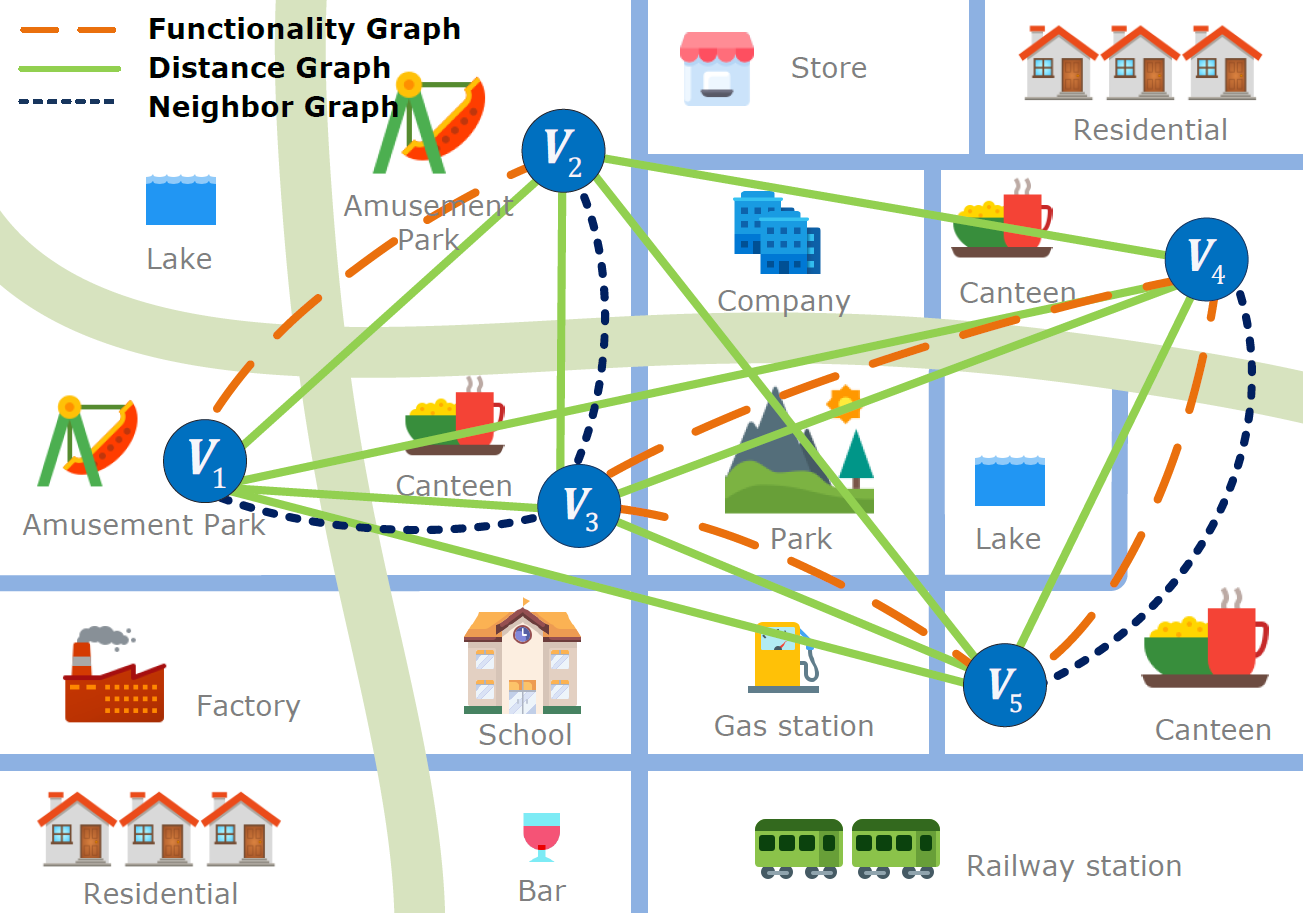} 
      \caption{Illustration of  multi-graph spatio-temporal forecasting.}
      \label{fig_MGNNs}
\end{figure}

One main challenge in LSTF is to effectively capture the long-term spatio-temporal dependency and extract contextual information. Recently, multi-graph neural networks (MGNNs) \cite{Wang2021Multigraph} have received increasing attention for spatio-temporal forecasting problems. Specifically, as shown in Figure~\ref{fig_MGNNs}, each node's value $V_i$ is estimated in the long run using historical data and correlations across nodes of a distance graph, where each edge denotes the correlation or dependency between two different nodes.
%Except for spatial correlation such as distance,
Furthermore, the functionality similarities of surrounding areas, which represent contextual information, can also be used for prediction purposes. Compared to the single graph approach, which may not comprehensively capture all the relationships, the MGNN-based approach is appropriate for leveraging more information and features by integrating different graphs. Thus, in this work, we choose the MGNN-based approach to infer how information about each node
evolves over time.

Although MGNNs show potential for extracting contextual information around prediction sites, four significant limitations remain when solving the LSTF problem:
\paragraph{(1) Most existing MGNN studies consider only the spatial similarity of nodes, such as the distance similarity and neighborhood correlation.} Previous studies have shown that the distance similarity is insufficient to represent correlations among nodes with spatio-temporal attributes \cite{geng2019spatiotemporal}. Wu \textit{et al.} \cite{wu2019graph} proposed an adaptive adjacency matrix to discover hidden spatial dependencies directly from historical records of each node in an end-to-end fashion by computing the inner product of the nodes' learnable embedding. However, these works did not utilize well the existing prior knowledge encoded as an adjacency matrix, which may result in missing vital information.
\paragraph{(2) Fusing different graph models is challenging.} For multi-graph-based problems, the graph models differ with different scales; thus, it is inappropriate to simply merge them using weighted sum or other averaging approaches. Additionally, how to align each node in different graphs is challenging since nodes in different graphs are associated with different spatio-temporal information.
\paragraph{(3) Existing multi-graph fusion approaches rely heavily on specific models.} The current MGNNs lack generalizability. Specifically, the existing graph construction approaches and fusion methods need to be strictly bonded, assuming specific graph neural network structures. 
Although such an end-to-end framework provides a convenient method, it induces various difficulties in examining the importance of each graph to find a better combination of each module.
\paragraph{(4) Long-term spatio-temporal dependency is not considered.} Usually, MGNNs tend to learn the spatio-temporal dependency by projecting mapping from data within the observation window and the prediction horizon. However, due to the limitation of data sources, existing graph models, such as the distance graph \cite{li2018diffusion} or the neighbor graph \cite{geng2019spatiotemporal} represent only the static spatial information, which cannot capture the long-term spatio-temporal dependency.

To address the issues above, we investigate graph construction and fusion mechanisms, and make improvements to each component. Specifically, we take advantage of human insights to build a new graph model namely `heuristic graph', which can represent the long-range distribution of the collected spatio-temporal data. Aiming to align various graphs with different information, we then employ the spatial and graph attention mechanisms to integrate nodes in the same graph and different graphs. Furthermore, to dynamically capture the contextual information and temporal dependency of each node in different graphs, we construct an adaptive correlation tensor to indicate the importance of each node.  
% The major advantage of spending efforts on new graphs construction and fusion rather than designing neural network components is as follows: first, graph constructions and fusion mechanisms can be used for deep learning neural networks and other learning algorithms. Second, we can observe the importance of each graph that contains specific information easily, which can also be used to further study the explainability of deep learning and improve models' transparency. 
In summary, the main contributions of this paper are as follows:

% \noindent \textbf{(1)} We propose a new graph model called a heuristic graph for the LSTF problem, which can represent the long-term spatio-temporal dependency from historical data or human insights, capture the long-term spatio-temporal dependency of the data, and can be widely used for various graph neural networks.
% \noindent \textbf{(2)} We design a novel graph model fusion module called a dynamic graph fusion block to integrate various graph models with graph attention and spatial attention mechanisms, aiming to align nodes within graphs and across different graphs. We further construct a learnable weight tensor for each node to flexibly capture the dynamic correlations between nodes.
% \noindent \textbf{(3)} We conduct extensive experiments on two large-scale public real-world spatio-temporal datasets. We validate the effectiveness of the proposed new graph models and fusion approaches using ablation studies. The code is available at {\color{blue}\underline{\url{https://github.com/swsamleo/HMSTGCN}}}.
\begin{itemize}
    \item We propose a new graph model namely `heuristic graph', for the LSTF problem, which can represent the long-term spatio-temporal dependency from historical data or human insights and can be widely used for various graph neural networks.
    \item We design a novel graph model fusion module called a dynamic graph fusion block to integrate various graph models with graph attention and spatial attention mechanisms, aiming to align nodes within graphs and across different graphs. We further construct a learnable weight tensor for each node to flexibly capture the dynamic correlations between nodes.
    \item We conduct extensive experiments on two large-scale public real-world spatio-temporal datasets. We validate the effectiveness of the proposed new graph models and fusion approaches using ablation studies. 
    %The code is available at {\color{blue}\underline{\url{https://github.com/swsamleo/MLSTGCN}}}.
\end{itemize}

\begin{figure*}[ht]
      \centering
      \includegraphics[width=0.9\textwidth]{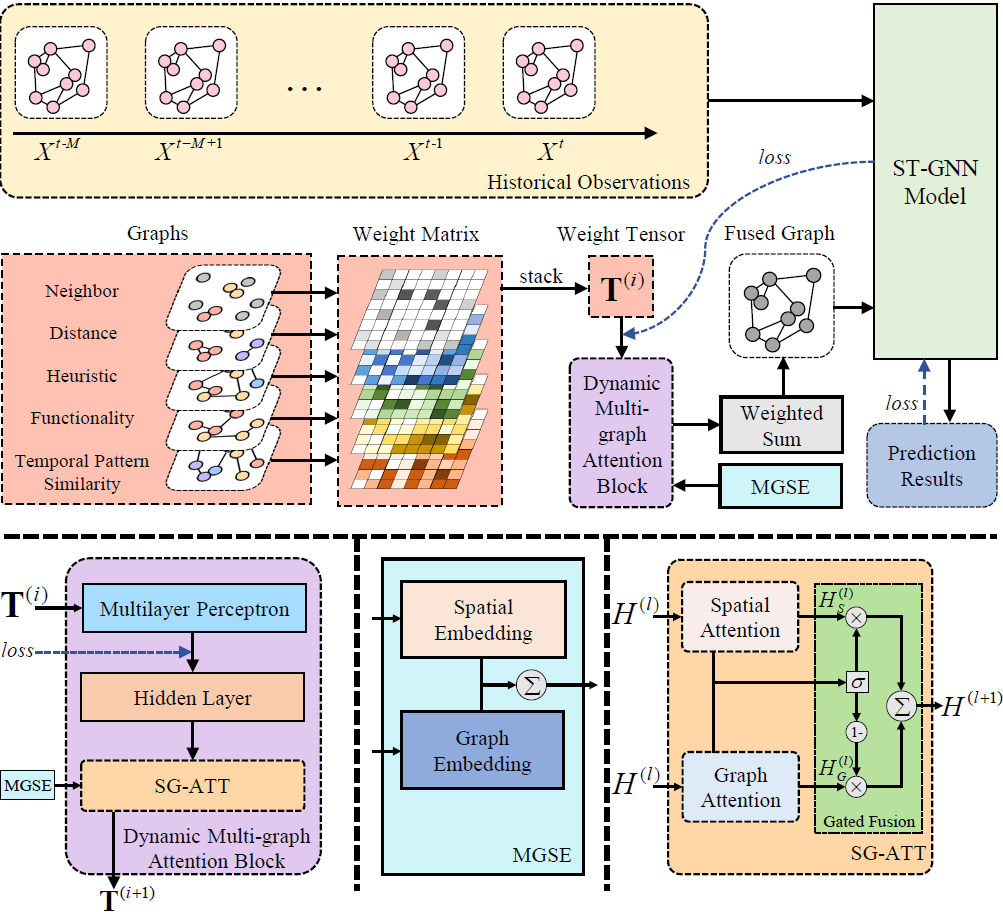} % Reduce the figure size so that it is slightly narrower than the column. Don't use precise values for figure width.This setup will avoid overfull boxes.
      \caption{The overview of the LSTF system.}
      \label{framework}
\end{figure*}

\section{Methodologies}
As shown in Figure \ref{framework}, the proposed framework consists of three major components: the graph construction module, the dynamic multi-graph fusion module, and the spatio-temporal graph neural network (ST-GNN). We designed five graphs to represent different aspects of the spatio-temporal information in the graph construction module. In the dynamic multi-graph fusion module, we align spatial and temporal dependency using an adaptive trainable tensor and introduce graph and spatial attention mechanisms to calculate the correlations among nodes located in different graphs. We then obtain the prediction results with existing ST-GNN models.
% which is the backend of the framework, and , which is used for spatio-temporal  in multi-graph. In the dynamic multi-graph fusion component, we construct five different graphs to extract contextual information, heuristic knowledge, temporal correlations, spatial correlations, and topology information. All details of implementation are shown in the following.

\subsection{Graph Construction}
% The graph is defined as $G=\{V,E,W\}$. 
In this section, we describe in detail two new graph models we proposed named the heuristic graph $G^H=\{V,E,W^H\}$ and the functionality graph $G^F=\{V,E,W^F\}$, combined with other three existing graphs, the distance graph $G^D=\{V,E,W^D\}$, neighbor graph $G^N=\{V,E,W^N\}$, and temporal pattern similarity graph $G^T=\{V,E,W^T\}$, into a multiple graph set $\mathbf{G} = \{G^D,G^N,G^F,G^H,G^T\}$.
%  We also propose a new attention-based graph fusion mechanism to merge all these different graphs into a single fusion graph $\mathcal{G}^*$.

%\subsubsection{Distance Graph}
\paragraph{Distance Graph.} The element of distance matrix $W^D$ is defined with a thresholded Gaussian kernel as follows \cite{shuman2013emerging}:
\begin{equation}
      \resizebox{.91\linewidth}{!}{$
      \displaystyle
      {W}^{D}_{ij}:= 
      \begin{cases}
            \exp \left(-\frac{d_{i j}^{2}}{\sigma_D^{2}}\right), & \text { for } i \neq j \text { and } \exp \left(-\frac{d_{i j}^{2}}{\sigma_D^{2}}\right) \geq \varepsilon, \\ 
            0, & \text { otherwise. }
      \end{cases}
      $}
\end{equation}
where $d_{ij}$ is the Euclidean distance between $v_i$ and $v_j$. $\varepsilon$ and $\sigma_D^2$ are used to control the sparsity and distribution of $W^D$.

%\subsubsection{Neighbor Graph}
\paragraph{Neighbor Graph.} The element of neighbor matrix $W^N$ is defined as follows:
\begin{equation}
      {W}^{N}_{ij}:=\begin{cases}1, & \text{if } v_i \text{ and } v_j\ \text{are adjacent},\\0 ,& \text{otherwise}.\end{cases}
\end{equation}

%\subsubsection{Functionality Graph}
% Functionality graph was first proposed by \cite{geng2019spatiotemporal}, the edge of which is defined as $E_{i, j}=\operatorname{sim}\left(P_{v_{i}}, P_{v_{j}}\right) \in[0,1]$, where $P_{v_{i}}$, $P_{v_{j}}$ are the functionality vectors of vertices $v_i$ and $v_j$, respectively and $\operatorname{sim}(\cdot)$ is used for calculating the similarity of $v_i$ and $v_j$. In this paper, we propose a new functionality graph using Pearson correlation coefficients because strong linear correlations exist between places with different functions, i.e., when two places have two similar functions, a strong correlation tends to exist within them. Specifically, each vertex has a series of functions, e.g, factories, schools, and hospitals. The total number of functions is $K$. The vector of the number of these functions of vector $v_i$ is denoted as $F_i=\{f_{i,1},f_{i,2},\cdots,f_{i,k} ,\cdots,f_{i,K} \}$. The functionality matrix then can be obtained using Pearson correlation coefficients \cite{schober2018correlation} by
\paragraph{Functionality Graph.} Usually, places with similar functionalities or utilities, such as factories, schools, and hospitals, have strong correlations. 
% In the real world, strong linear correlations exist between places with different functions, i.e., when two places have two similar functions such as factories, schools, and hospitals, a strong correlation tends to exist within them. 
In this paper, different from the functionality graph proposed by \cite{geng2019spatiotemporal}, we propose a new functionality graph using Pearson correlation coefficients to capture the global contextual function similarity. Denote the total number of functions is $K$; then the vector of the number of these functions of vertex $v_i$ is denoted as $F_i=\{f_{i,1},f_{i,2},\cdots,f_{i,k} ,\cdots,f_{i,K} \}$. The functionality matrix can be obtained using Pearson correlation coefficients \cite{zhang2020crowd} by
\begin{equation}
      \resizebox{.91\linewidth}{!}{$
      \displaystyle
        W^{F}_{i j}:= 
        \begin{cases}\frac{\sum_{k=1}^{K}\left(f_{i, k}-\overline{F_{i}}\right)\left(f_{j, k}-\overline{F_{j}}\right)}{\sqrt{\sum_{i=1}^{k}\left(f_{i, k}-\overline{F_{i}}\right)^{2}} \sqrt{\sum_{j=1}^{k}\left(f_{j, k}-\overline{F_{j}}\right)^{2}}}, & \text { if } i \neq j, \\
            0, & \text{otherwise.}
      \end{cases}
    $}
\end{equation}
% Constructing the graph via using Pearson correlation coefficients, we can better capture the global contextual function similarity. 
Note that we consider all functions that contribute equally to the relationships of nodes.

%\subsubsection{Heuristic Graph}
\paragraph{Heuristic Graph.} To leverage heuristic knowledge and human insights, we propose a new graph model called the heuristic graph. We create a histogram to represent the overview of the spatio-temporal training data, where each bin indicates a predefined temporal range, and the bar height measures the number of data records that fall into each bin. Then we apply a function $f(x) = \alpha e^{-\beta x}$ to approximate the histogram. For a vertex $v_i$, we can obtain its fitted parameters $\alpha_i$ and $\beta_i$. The distribution distance is calculated using the Euclidean distance ${d}^{H}_{ij} = \sqrt{(\alpha_1 - \alpha_2)^2 + (\beta_1 - \beta_2)^2}$. The element of the heuristic matrix $W^H$ can be defined as follows:
\begin{equation}
      {W}^{H}_{ij}:=\begin{cases}\text{exp}\left( -\frac{\|{d}^{H}_{{ij}}\|^2}{\sigma_H^2}\right), & \text{for } i \ne j,\\0 ,& \text{otherwise}.\end{cases}
\end{equation}
where $\sigma_H^2$ is a parameter to control the distribution of $W^H$. Kullback-Leibler (KL) divergence \cite{van2014renyi} can be also used to create this graph, which usually quantifies the difference between two probability distributions.

%\subsubsection{Temporal Pattern Similarity Graph}
\paragraph{Temporal Pattern Similarity Graph.} For a vertex $v_{i}$, the vector of the time-series data used for training is described as $T_i = \{t_{i,1},t_{i,2},\cdots,t_{i,p} ,\cdots,t_{i,P} \}$, where $P$ is the length of the series, and $t_{i,p}$ is the time-series data value of the vertex $v_i$ at time step $p$. 
We also use the Pearson correlation coefficients \cite{zhang2020crowd} to define the elements of the temporal pattern similarity matrix $W^T$ as follows:
\begin{equation}
      \resizebox{.91\linewidth}{!}{$
      \displaystyle
        W^{T}_{i j}:=
        \begin{cases}
            \frac{\sum_{p=1}^{P}\left(t_{i, p}-\overline{T_{i}}\right)\left(t_{j, p}-\overline{T_{j}}\right)}{\sqrt{\sum_{i=1}^{p}\left(t_{i, p}-\overline{T_{i}}\right)^{2}} \sqrt{\sum_{j=1}^{p}\left(f_{j, p}-\overline{T_{j}}\right)^{2}}}, & \text { if } i \neq j, \\
            0, & \text{otherwise.}
        \end{cases}
    $}
\end{equation}

\subsection{Dynamic Multi-graph Fusion}

\begin{algorithm}[tb]
      \caption{Dynamic Multi-graph Fusion}
      \label{alg:algorithm}
      \textbf{Input}: Weight matrices: $W^D$, $W^N$, $W^F$, $W^H$, $W^T$\\
      \textbf{Parameter}: Number of batches: $\mathrm{Bt}$\\
      \textbf{Output}: Fused weight matrix $W^*$
      \begin{algorithmic}[1] %[1] enables line numbers
      \STATE Stack weight matrices to tensor $\mathbf{T}^{(0)} \in \mathbb{R}^{|\mathbf{G}| \times N \times N}$.
      \STATE Train $\mathbf{T}^{(0)}$ while training ST-GNN models.
      \FOR{$i \in [0,\mathrm{Bt}-1]$}
      \STATE $\mathbf{T}^{(i+1)} \longleftarrow \mathbf{T}^{(i)}$
      \STATE $i=i+1$
      \STATE $W^{*}_{jk} = \sum^{|\mathbf{G}|}_{i=1}\mathbf{T}^{(\mathrm{i})}(i,j,k)$, where $W^{*}_{jk}$ is the element of the weight matrix of the fused graph.
      \ENDFOR
      \STATE \textbf{return} $W^{*}$
      \end{algorithmic}
\end{algorithm}

The graph fusion approach plays a key role in multi-graph neural networks as multi-graphs cannot simply be merged with the average sum or the weighted sum \cite{wang2020multi}. In this paper, a dynamic graph fusion method is proposed; the whole process of this method is shown in Figure~\ref{framework} and Algorithm~\ref{alg:algorithm}. We construct a trainable weight tensor as the input of a dynamic multi-graph attention block (DMGAB). Moreover, we incorporate the spatial and graph information into multi-graph spatial embedding (MGSE) and add this embedding to the DMGAB. To facilitate the residual connection, all layers of the DMGAB produce outputs of $D$ dimensions, and the block can be expressed as $\mathrm{DMGAB} \in \mathbb{R}^{|\mathbf{G}| \times N \times D}$.
% Since the weight tensor is trainable, after obtaining the L1 loss for each epoch, the weight coefficients of each graph or each node of the graph are adaptively adjusted based on the loss backward. A new weight tensor is created and the training process is looped.

\subsubsection{Multi-graph Spatial Embedding} 
We apply the spatial embedding $E_{v_i}^S \in \mathbb{R}^D$ to preserve the graph structure information. To represent the relationships of the nodes in different graphs, we further propose graph embedding to encode five graphs into $\mathbb{R}^{|\mathbf{G}|}$. Then we employ a two-layer fully-connected neural network to transform the graphs into a vector $\mathbb{R}^D$ and obtain the multi-graph embedding $E_{G_i}^{MG} \in \mathbb{R}^{D}$, where $G_i$ is any graph. To obtain the vertex representations among multiple graphs, we fuse the spatial embedding and the multi-graph embedding as the multi-graph spatial embedding (MGSE) with $E_{v_i,G_i} = E_{v_i}^S + E_{G_i}^{MG}$.

\begin{figure}[t]
      \centering
            \begin{subfigure}{0.47\textwidth} % width of left subfigure
                  \includegraphics[width=\textwidth]{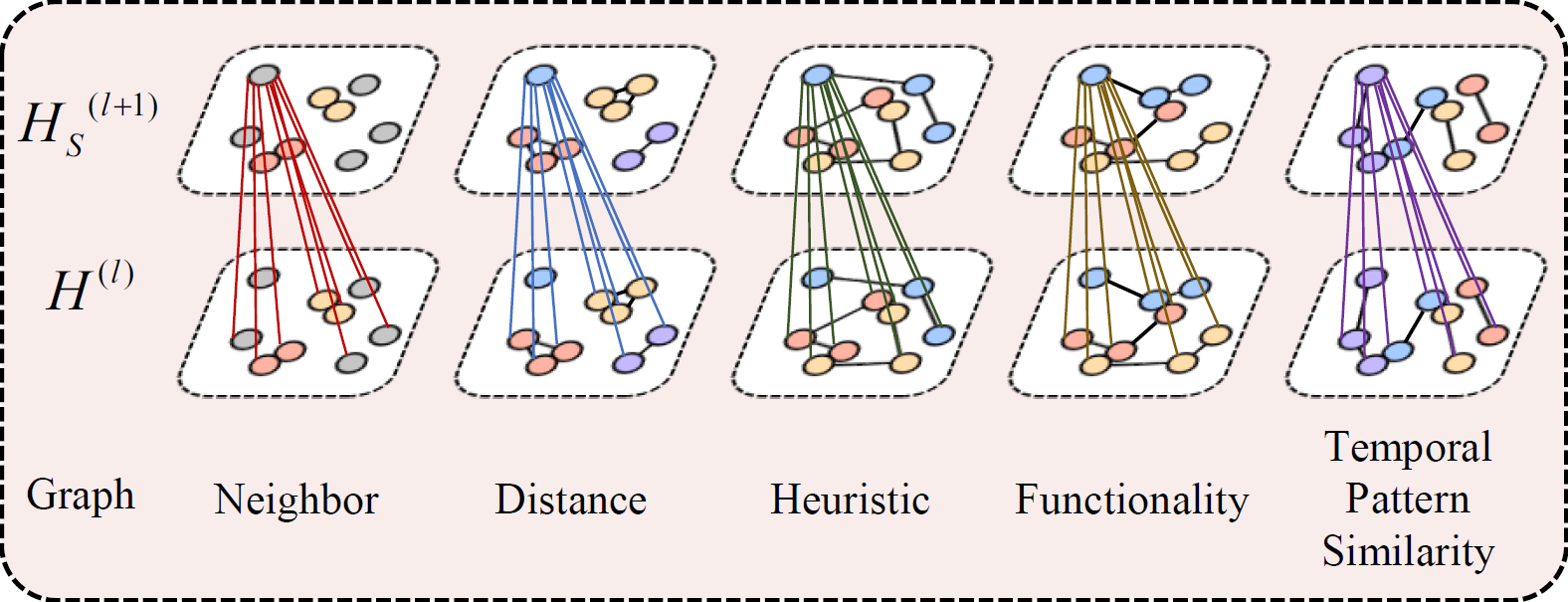}
                  \caption{spatial attention} % subcaption
                  \label{spatial attention}
            \end{subfigure}
            \begin{subfigure}{0.47\textwidth} % width of left 
                  \includegraphics[width=\textwidth]{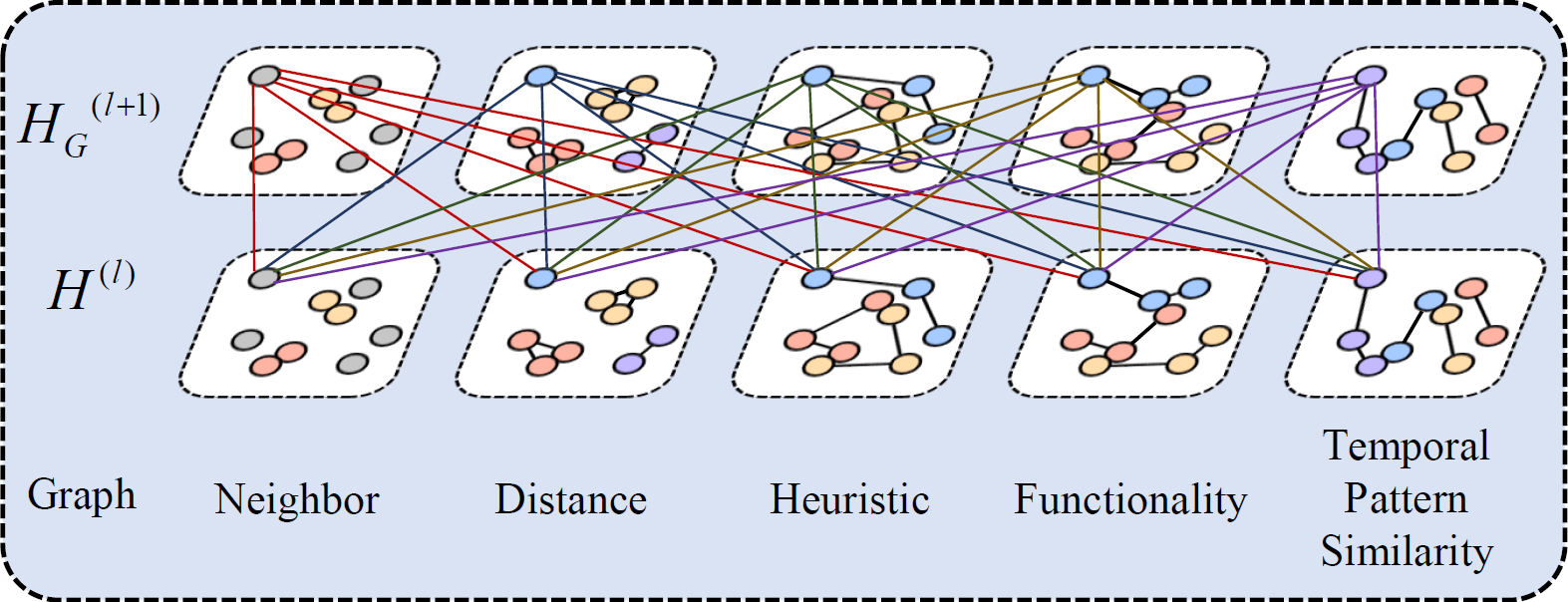}
                  \caption{graph attention}
                  \label{graph attention}
            \end{subfigure}
            \caption{The attention mechanisms adopted in this paper.} % caption for whole figure
\end{figure}

\subsubsection{Dynamic Multi-graph Attention Block}
Any node in a graph is impacted by other nodes with different levels. When acting on multiple graphs, these impacts are magnified. To model inner node correlations, we design a multi-graph attention block to adaptively capture the correlations among the nodes. 
% In the dynamic multi-graph attention block, the weight tensor is passed into an MLP to make the weights trainable. 
As shown in Figure~\ref{framework}, the multi-graph attention block contains spatial attention and graph attention. We denote the input of the $l$-th block $H^{(l)}$ and denote the hidden state of the vertex $v_i$ on graph $G_i$ in $H^{(l)}$ as $h_{v_i,G_i}^{(l)}$. The output blocks of the spatial and graph attention mechanisms are denoted as $H_S^{(l+1)}$ and $H_G^{(l+1)}$, respectively.

\paragraph{Spatial Attention.} Inspired by \cite{zheng2020gman}, we capture the contextual correlations of nodes by proposing a spatial attention mechanism (shown in Figure~\ref{spatial attention}). Different from the previous spatial attention mechanism, which acts on the hidden state of the batch of temporal data, our method acts on the hidden state of the weight tensor. Then we can calculate the next hidden state in the graph $G_i$ as follows:
\begin{equation}
      hs_{v_{i}, G_i}^{(l+1)}=\sum_{v_k \in \mathcal{V}_i} \alpha_{v_{i}, v_k} \cdot h_{v_k, G_i}^{(l)}
\end{equation}
where $\mathcal{V}_i$ is all the vertices on the graph except the $v_i$. $\alpha_{v_i,v_k}$ is the attention score respecting the importance of $v_k$ to $v_i$. 

In the real world, the vertices are influenced not only by other vertices on the same graph but also other graphs. For example, the parking occupancy rate of one place is affected not only by the distance from another place but also by the functionality of another place. To this end, we concatenate the hidden state with MGSE to extract both the spatial and graph features and employ the scaled dot-product approach to calculate the relevance between $v_i$ and $v_k$ with 
\begin{equation}
      s_{v_{i}, v_{k}}=\frac{\left\langle h_{v_{i}, G_{i}}^{(l)}\left\|E_{v_{i}, G_{i}}, h_{v_k, G_{i}}^{(l)}\right\| E_{v_k, G_{i}}\right\rangle}{\sqrt{2 D}},
\end{equation}
where $\|$ is the concatenation operation and $\langle \cdot|\cdot \rangle$ is the inner product operation. Then $s_{v_{i}, v_{k}}$ is normalized by the softmax function $\alpha_{v_{i}, v_k}=\exp \left(s_{v_{i}, v_k}\right)/\sum_{v_k \in \mathcal{V}_i} \exp \left(s_{v_{i}, v_{k}}\right)$.
% \begin{equation}
%       \alpha_{v_{i}, v_k}=\frac{\exp \left(s_{v_{i}, v_k}\right)}{\sum_{n=1}^{N}  \exp \left(s_{v_{i}, v_{n}}\right)}.
%       \label{alpha ij}
% \end{equation}
To stabilize the learning process, we concatenate $M$ parallel attention mechanisms to extend them to the multi-head attention mechanism \cite{zheng2020gman} with
\begin{equation}
      \resizebox{.91\linewidth}{!}{$
      \displaystyle
      s_{v_i, v_k}^{(m)}=\frac{\left\langle f_{s, 1}^{(m)}\left(h_{v_{i}, G_{i}}^{(l)} \| E_{v_{i}, G_{i}}\right), f_{s, 2}^{(m)}\left(h_{v_k, G_{i}}^{(l)} \| E_{v_k, G_{i}}\right)\right\rangle}{\sqrt{d}},
      $}
\end{equation}
\begin{equation}
      % \alpha_{v_i, v_k}^{(m)}&=\frac{\exp \left(s_{v_{i}, v_k}^{(m)}\right)}{\sum_{n=1}^{N}  \exp \left(s_{v_{i}, v_{n}}^{(m)}\right)} \\
      h s_{v_{i}, G_{i}}^{(l+1)}=\|_{m=1}^{M}\left\{\sum_{n=1}^{N}\alpha_{v_{i}, v_n}^{(m)} \cdot f_{s, 3}^{(m)}\left(h_{v_n, G_{i}}^{(l)}\right)\right\},
\end{equation}
where $f_{s, 1}^{(m)}(\cdot)$, $f_{s, 2}^{(m)}(\cdot)$, and $f_{s, 3}^{(m)}(\cdot)$ are different $\mathrm{ReLU}$ functions serving as nonlinear projections in $m$-th head attention. $\alpha_{v_{i}, v_n}^{(m)}$ is calculated with a softmax function in the $m$-th head attention and $h s_{v_{i}, G_{i}}^{(l+1)}$ is the hidden state of $v_i \in G_i$.

\begin{table*}[ht]
      \centering
      \scalebox{0.9}{
        %\begin{tabular}{ccc||cc||cc||cc||cc||cc||cc}
      \begin{tabular}{ccc||p{22pt}<{\centering}p{36pt}<{\centering}||p{32pt}<{\centering}p{44pt}<{\centering}||p{28pt}<{\centering}p{40pt}<{\centering}||p{28pt}<{\centering}p{40pt}<{\centering}||p{28pt}<{\centering}p{32pt}<{\centering}}
            \toprule
            \multicolumn{1}{c|}{\multirow{2}[2]{*}{\small{Datasets}}} & \multicolumn{2}{c||}{\small{Methods}} & \multirow{2}[2]{*}{\small{STGCN}} & \multirow{2}[2]{*}{\small{STGCN*}} & \multirow{2}[2]{*}{\small{ST-MGCN}} & \multirow{2}[2]{*}{\small{ST-MGCN*}} & \multirow{2}[2]{*}{\small{ASTGCN}} & \multirow{2}[2]{*}{\small{ASTGCN*}} & \multirow{2}[2]{*}{\small{MSTGCN}} & \multirow{2}[2]{*}{\small{MSTGCN*}} & \multicolumn{1}{c}{\multirow{2}[2]{*}{\makecell[c]{\small{Graph}\\\small{WaveNet}}}} & \multicolumn{1}{c}{\multirow{2}[2]{*}{\makecell[c]{\small{Graph}\\\small{WaveNet*}}}} \\
            \multicolumn{1}{c|}{} & \multicolumn{2}{c||}{\small{Metric}} &       &       &       &       &       &       &       &       &       &  \\
            \specialrule{.08em}{.4ex}{.65ex}
            \multicolumn{1}{c|}{\multirow{16}[4]{*}{\begin{sideways}Parking\end{sideways}}} & \multicolumn{1}{c|}{\multirow{8}[2]{*}{\begin{sideways}RMSE\end{sideways}}} & 3     & 0.0607  & \textbf{0.0514} & 0.0553  & \textbf{0.0524} & 0.0517  & \textbf{0.0493}& 0.0604  & \textbf{0.0479}& 0.0477& \textbf{0.0473}  \\
            \multicolumn{1}{c|}{} & \multicolumn{1}{c|}{} & 6     & 0.0751  & \textbf{0.0658}& 0.0677  & \textbf{0.0646}& 0.0642  & \textbf{0.0611}& 0.0724  & \textbf{0.0607}  & 0.0608 & \textbf{0.0594}  \\
            \multicolumn{1}{c|}{} & \multicolumn{1}{c|}{} & 9     & 0.0869  & \textbf{0.0787}& 0.0794  & \textbf{0.0748}& 0.0748  & \textbf{0.0702}& 0.0833  & \textbf{0.0706}  & 0.0709 & \textbf{0.0684}  \\
            \multicolumn{1}{c|}{} & \multicolumn{1}{c|}{} & 12    & 0.0992  & \textbf{0.0903}& 0.0900  & \textbf{0.0839}& 0.0843  & \textbf{0.0776}& 0.0939  & \textbf{0.0796}& 0.0801  & \textbf{0.0762}\\
            \multicolumn{1}{c|}{} & \multicolumn{1}{c|}{} & 15    & 0.1107  & \textbf{0.1007}& 0.0999  & \textbf{0.0924}& 0.1115  & \textbf{0.0850}& 0.1041  & \textbf{0.0875}& 0.0883  & \textbf{0.0832}\\
            \multicolumn{1}{c|}{} & \multicolumn{1}{c|}{} & 18    & 0.1210  & \textbf{0.1105}& 0.1092  & \textbf{0.1002}& 0.1228  & \textbf{0.0915}& 0.1140  & \textbf{0.0947}& 0.0965  & \textbf{0.0893}\\
            \multicolumn{1}{c|}{} & \multicolumn{1}{c|}{} & 21    & 0.1301  & \textbf{0.1194}& 0.1178  & \textbf{0.1075}& 0.1308  & \textbf{0.0977}& 0.1229  & \textbf{0.1017}& 0.1040  & \textbf{0.0946}\\
            \multicolumn{1}{c|}{} & \multicolumn{1}{c|}{} & 24    & 0.1393  & \textbf{0.1276}& 0.1259  & \textbf{0.1143}& 0.1410  & \textbf{0.1035}& 0.1318  & \textbf{0.1071}& 0.1114  & \textbf{0.0996}\\
        \cmidrule{2-13}    \multicolumn{1}{c|}{} & \multicolumn{1}{c|}{\multirow{8}[2]{*}{\begin{sideways}MAE\end{sideways}}} & 3     & 0.0425 & \textbf{0.0358}  & 0.0376  & \textbf{0.0360}& 0.0362  & \textbf{0.0342}& 0.0443  & \textbf{0.0328} & 0.0323 & \textbf{0.0321}  \\
            \multicolumn{1}{c|}{} & \multicolumn{1}{c|}{} & 6     & 0.0529  & \textbf{0.0457}& 0.0467  & \textbf{0.0450}& 0.0452  & \textbf{0.0431}& 0.0532  & \textbf{0.0424}  & 0.0422 & \textbf{0.0416}  \\
            \multicolumn{1}{c|}{} & \multicolumn{1}{c|}{} & 9     & 0.0616  & \textbf{0.0555}& 0.0556  & \textbf{0.0529}& 0.0531  & \textbf{0.0504}& 0.0613  & \textbf{0.0502}  & 0.0499 & \textbf{0.0485}  \\
            \multicolumn{1}{c|}{} & \multicolumn{1}{c|}{} & 12    & 0.0711  & \textbf{0.0640}& 0.0638  & \textbf{0.0600}& 0.0601  & \textbf{0.0565}& 0.0692  & \textbf{0.0575}  & 0.0571 & \textbf{0.0545}  \\
            \multicolumn{1}{c|}{} & \multicolumn{1}{c|}{} & 15    & 0.0798  & \textbf{0.0719}& 0.0715  & \textbf{0.0667}& 0.0675  & \textbf{0.0623}& 0.0769  & \textbf{0.0637}& 0.0635  & \textbf{0.0599}\\
            \multicolumn{1}{c|}{} & \multicolumn{1}{c|}{} & 18    & 0.0882  & \textbf{0.0795}& 0.0788  & \textbf{0.0728}& 0.0742  & \textbf{0.0674}& 0.0846  & \textbf{0.0695}& 0.0698  & \textbf{0.0646}\\
            \multicolumn{1}{c|}{} & \multicolumn{1}{c|}{} & 21    & 0.0956  & \textbf{0.0866}& 0.0855  & \textbf{0.0786}& 0.0810  & \textbf{0.0722}& 0.0918  & \textbf{0.0752}& 0.0755  & \textbf{0.0687}\\
            \multicolumn{1}{c|}{} & \multicolumn{1}{c|}{} & 24    & 0.1036  & \textbf{0.0934}& 0.0919  & \textbf{0.0841}& 0.0867  & \textbf{0.0768}& 0.0995  & \textbf{0.0794}& 0.0812  & \textbf{0.0724}\\
            \specialrule{.08em}{.4ex}{.65ex}
            \multicolumn{1}{c|}{\multirow{16}[4]{*}{\begin{sideways}Air Quality\end{sideways}}} & \multicolumn{1}{c|}{\multirow{8}[2]{*}{\begin{sideways}RMSE\end{sideways}}} & 3     & 6.6843  & \textbf{6.3609}& 6.7802  & \textbf{6.3255}& \textbf{6.3958}  & 6.9846 & 7.0427  & \textbf{6.3729}& 6.4878 & \textbf{6.1353}  \\
            \multicolumn{1}{c|}{} & \multicolumn{1}{c|}{} & 6     & 8.3989  & \textbf{7.7995}& 8.7083  & \textbf{7.6470}& \textbf{7.7519}  & 8.1628 & 8.4449  & \textbf{7.6549}& \textbf{8.2323}  & 8.5556\\
            \multicolumn{1}{c|}{} & \multicolumn{1}{c|}{} & 9     & 9.7762  & \textbf{8.6881}& 10.3522  & \textbf{8.6431}& \textbf{9.0522}  & 9.4715 & 9.3188  & \textbf{8.7717}& \textbf{10.3232}  & 10.8160\\
            \multicolumn{1}{c|}{} & \multicolumn{1}{c|}{} & 12    & 10.8079  & \textbf{9.5392}& 11.5615  & \textbf{9.5453} & 10.7794  &\textbf{10.2963} & 10.7145  & \textbf{9.6747}& \textbf{12.9487}  & 13.1379\\
            \multicolumn{1}{c|}{} & \multicolumn{1}{c|}{} & 15    & 11.7172  & \textbf{10.1575}& 12.3340  & \textbf{10.3465} & 11.9669 &\textbf{10.9218}  & 11.4235  & \textbf{10.7134}& 15.7093  & \textbf{15.0418}\\
            \multicolumn{1}{c|}{} & \multicolumn{1}{c|}{} & 18    & 11.9014  & \textbf{10.4241}& 12.7944  & \textbf{10.9299} & 13.2015 & \textbf{11.8600}  & 12.3950  & \textbf{11.1146}& 19.2235  & \textbf{14.1381}\\
            \multicolumn{1}{c|}{} & \multicolumn{1}{c|}{} & 21    & 12.5268  & \textbf{11.3408}& 13.1333  & \textbf{11.2794} & 14.4416 & \textbf{11.6768}  & 13.1675  & \textbf{10.7613}& 21.1240  & \textbf{13.4125}\\
            \multicolumn{1}{c|}{} & \multicolumn{1}{c|}{} & 24    & 12.9587  & \textbf{11.8283}& 13.4853  & \textbf{11.3442} & 14.6537 & \textbf{10.7624}  & 13.3226  & \textbf{11.3835}& 21.2758  & \textbf{13.2053}\\
        \cmidrule{2-13}    \multicolumn{1}{c|}{} & \multicolumn{1}{c|}{\multirow{8}[2]{*}{\begin{sideways}MAE\end{sideways}}} & 3     & 4.8973  & \textbf{4.5597}& 5.2823  & \textbf{4.5859}& \textbf{4.7643}  & 5.4481& 5.3349  & \textbf{4.7283}& 4.9826 & \textbf{4.4299}  \\
            \multicolumn{1}{c|}{} & \multicolumn{1}{c|}{} & 6     & 6.9107  & \textbf{6.0212}& 7.1787  & \textbf{5.7940}& \textbf{6.0111}  & 6.5127& 6.7291  & \textbf{5.9757}& \textbf{6.6243}  & 6.8427\\
            \multicolumn{1}{c|}{} & \multicolumn{1}{c|}{} & 9     & 7.7049  & \textbf{6.9780}& 8.8038  & \textbf{6.7984}& \textbf{7.2884}  & 7.8179& 7.6316  & \textbf{7.0500}& \textbf{8.6868}  & 8.9678\\
            \multicolumn{1}{c|}{} & \multicolumn{1}{c|}{} & 12    & 8.8756  & \textbf{7.9221}& 9.9872  & \textbf{7.7235}& 9.1761  & \textbf{8.6078}& 9.1154  & \textbf{7.9514}& 11.1945  & \textbf{11.0713}\\
            \multicolumn{1}{c|}{} & \multicolumn{1}{c|}{} & 15    & 9.3082  & \textbf{8.5626}& 10.7280  & \textbf{8.5209}& 10.3348 & \textbf{9.1881}  & 9.8296  & \textbf{9.0196}& 13.6732  & \textbf{12.3349}\\
            \multicolumn{1}{c|}{} & \multicolumn{1}{c|}{} & 18    & 10.2937  & \textbf{8.8195}& 11.1683  & \textbf{9.0835}& 11.5575 & \textbf{9.0804}  & 10.7857  & \textbf{9.3868}& 16.5516  & \textbf{11.8929}\\
            \multicolumn{1}{c|}{} & \multicolumn{1}{c|}{} & 21    & 10.7669  & \textbf{9.7565}& 11.4962  & \textbf{9.4271}& 12.7541 & \textbf{8.8657}  & 11.5581  & \textbf{8.9675}& 18.0054  & \textbf{11.1300}\\
            \multicolumn{1}{c|}{} & \multicolumn{1}{c|}{} & 24    & 10.9152  & \textbf{10.2256}& 11.8230  & \textbf{9.5085}& 12.9317 & \textbf{8.9076}  & 11.6787  & \textbf{9.6138}& 18.1966  & \textbf{10.8037}\\
            \specialrule{.08em}{.4ex}{.65ex}
            \multicolumn{3}{c||}{Count} & 0     & 32    & 0     & 32    & 6     & 26    & 0     & 32    & 5     & 27 \\
            \bottomrule
      \end{tabular}%
      }
      \caption{The prediction results with five ST-GNN models with or without multi-graph modules on two datasets. (`*' indicates the ST-GNN model with the proposed dynamic multi-graph fusion method.)}
      \label{tab:X1V1}%
\end{table*}%

\paragraph{Graph Attention.} We employ graph attention to obtain the self-correlations of a node in different graphs (as shown in Figure~\ref{graph attention}). Similar to the spatial attention mechanism, we concatenate the hidden state with MGSE and employ the multi-head method to calculate the correlations. For $v_i$, the correlation between graph $G_j$ and $G_k$ is defined as:

\begin{equation}
      \resizebox{.89\linewidth}{!}{$
      \displaystyle
      u_{G_{j},G_k}^{(m)}=\frac{\left\langle f_{G, 1}^{(m)}\left(h_{v_{i}, G_{j}}^{(l)} \| E_{v_{i}, G_{j}}\right), f_{G, 2}^{(m)}\left(h_{v_{i}, G_k}^{(l)} \| E_{v_{i}, G_k}\right)\right\rangle}{\sqrt{d}},
      $}
\end{equation}
\begin{equation}
      % \beta_{g_{j}, g_k}^{(m)}=&\frac{\exp \left(u_{g_{j}, g_k}^{(m)}\right)}{\sum_{a=1}^{A} \exp \left(u_{g_{j}, g_{a}}^{(m)}\right)},\\
      h g_{v_{i}, G_{j}}^{(l+1)}=\|_{m=1}^{M}\left\{\sum_{k=1}^{|\mathbf{G}|}\beta_{G_{j}, G_k}^{(m)} \cdot f_{G, 3}^{(m)}\left(h_{v_i, G_{k}}^{(l)}\right)\right\},
\end{equation}
where $\beta_{G_{j}, G_k}^{(m)}$ calculated with a softmax function is the attention score in the $m$-th head, indicating the importance of graph $G_k$ to $G_j$, $f_{G, 1}^{(m)}(\cdot)$, $f_{G, 2}^{(m)}(\cdot)$, and $f_{G, 3}^{(m)}(\cdot)$ are the $\mathrm{ReLU}$ functions in $m$-th head attention.

\paragraph{Gated Fusion.} To further extract the correlations of nodes on different graphs, 
% Since the feature of one node is affected by other nodes in the same graph and the node on other graphs, 
we adopt the gated fusion method \cite{zheng2020gman} to consider both effects. The spatial attention $H_S^{(l)}$ and the graph attention $H_G^{(l)}$ in the $l$-th block are fused with
\begin{equation}
      H^{(l)}=z \odot H_{S}^{(l)}+(1-z) \odot H_{G}^{(l)},
\end{equation}
where the gate $z$ is calculated by:
\begin{equation}
      z=\sigma\left(H_{S}^{(l)} \mathbf{W}_{z, 1}+H_{G}^{(l)} \mathbf{W}_{z, 2}+\mathbf{b}_{z}\right),
\end{equation}
where $\mathbf{W}_{z, 1} \in \mathbb{R}^{D \times D}$, $\mathbf{W}_{z, 2} \in \mathbb{R}^{D \times D}$, and $\mathbf{b}_{z} \in \mathbb{R}^{D}$ are the learnable parameters, $\odot$ indicates the element-wise Hadamard product, and $\sigma(\cdot)$ is the $\mathrm{sigmoid}$ activation function. By combining the spatial and graph attention mechanisms, we further create a spatial-graph attention (SG-ATT) block, which is shown in Figure~\ref{framework}.

\section{Experiments}
\subsection{Datasets}
% We extensively conduct experiments on two large-scale public datasets which are present as follows:
% The two datasets used in this paper are listed as follows:

\noindent\textit{Parking: }The Melbourne parking dataset\footnote{\url{https://data.melbourne.vic.gov.au/}}, collected by the Melbourne City Council in 2019, contains $42,672,743$ parking events recorded by the in-ground sensors every five minutes located in the Melbourne Central Business District (CBD) \cite{shao2017traveling}. All sensors have been classified into 40 areas. 
% To obtain the functionality of each area, the Point of Interest (PoI) data, which recorded business establishments in 2019, was also collected.
% The PoI data\footnote{\url{https://data.melbourne.vic.gov.au/Business/Business-establishment-trading-name-and-industry-c/vesm-c7r2}} contains 381 functions, including restaurants, higher education, etc. 

\noindent\textit{Air Quality: }The Ministry of Ecology and Environment of China (MEE)\footnote{\url{https://english.mee.gov.cn/}} published a large-scale air quality dataset \cite{wang2020pm25}, comprising 92 air quality monitoring stations, to assess the hourly $\mathrm{PM}_{2.5}$ concentration in Jiangsu province in 2020. 
% The missing data were replaced by the latest valid observation following \cite{gclstm}. 

\subsection{Experimental Details}
\paragraph{Baselines.} We selected five state-of-the-art ST-GNN models as baselines: STGCN \cite{yu_spatio-temporal_2018}, ASTGCN \cite{guo2019attention}, MSTGCN \cite{guo2019attention}, ST-MGCN \cite{geng2019spatiotemporal}, and Graph WaveNet \cite{wu2019graph}.
% \begin{itemize}
%     \item \textbf{STGCN:} The Spatio-temporal Graph Convolutional Network is an end-to-end deep learning neural network that combines 1-D CNN with gated linear units (GLUs) and graph convolutional networks to capture spatio-temporal correlations \cite{yu_spatio-temporal_2018}.
%     \item \textbf{ASTGCN:} The Attention-based spatio-temporal Graph Convolution Network \cite{guo2019attention} adds the Spatio-temporal attention mechanism to the spatio-temporal convolution to capture the dynamic spatio-temporal correlations.
%     \item \textbf{MSTGCN:} The Multi-Component spatio-temporal Graph Convolution Networks  \cite{guo2019attention} have the same settings as ASTGCN without the spatio-temporal attention mechanism.
%     \item \textbf{ST-MGCN:} The Spatio-temporal Multi-graph Convolution Network \cite{geng2019spatiotemporal} applies multi-graph convolution to modeling non-Euclidean correlations among regions and employs Contextual Gated RNN (CGRNN) to capture temporal dependencies.
%     \item \textbf{Graph WaveNet:} The Graph WaveNet uses a self-adaptive adjacency matrix to preserve hidden spatial dependencies and assembles the graph convolution with dilated casual convolution to capture spatio-temporal dependencies \cite{wu2019graph}.
% \end{itemize}

\paragraph{Platform.} All experiments were trained and tested on a Linux system (CPU: Intel(R) Xeon(R) Gold 6240 CPU @2.60GHz, GPU: NVIDIA GeForce RTX 2080 Ti).

\paragraph{Hyper-parameters.} All the tests used a 24-time step historical time window, and the prediction horizons ranged from three to 24 steps. 
% Specifically, we used 24 observed data points to predict the parking occupancy rate or the concentrations of the $\mathrm{PM}_{2.5}$ in the next 24 time steps. 
The proposed methods were optimized with the Adam optimizer. The learning rate was set to $1 e^{-4}$. The L1 loss function was adopted to measure the performance of the proposed model. The batch size was 32, and the global seed was set to 0 for the experiment repeat. All the tests were trained for 40 epoches. The number of attention heads $M$ and the dimension $d$ of each attention head were set to 8 and 8 in the \textit{Parking} dataset and set to 24 and 6 in the \textit{Air Quality} dataset.

\paragraph{Evaluation Metrics.} In our study, mean absolute error (MAE) and root mean square error (RMSE) were used.
% \begin{itemize}
%       \item \textbf{Platform:} All experiments were trained and tested on a Linux system (CPU: Intel(R) Xeon(R) Gold 6240 CPU @2.60GHz, GPU: NVIDIA GeForce RTX 2080 Ti). 
%     %   The source code is available at \url{https://github.com/swsamleo/HMSTGCN}.
%       \item \textbf{Hyper-parameters:} All the tests used a 24-time step historical time window, and the prediction horizons ranged from 3 to 24 steps. Specifically, we used 24 observed data points ($M$ = 24) to predict the parking occupancy rate or the concentrations of the $\mathrm{PM}_{2.5}$ in the next 24 time steps ($H$ = 24). Our proposed methods were optimized with Adam optimizer. The learning rate was set to $1 e^{-4}$. The L1 loss function was adopted to measure the performance of the proposed model. The batch size in all experiments was 32, and the global seed was set to 0 for experiments repeat. The number of attention heads $M$, and the dimension $d$ of each attention head are set to 8, 8 in \textbf{Parking} dataset and set to 24, 8 in \textbf{Air Quality} dataset.
%       \item \textbf{Evaluation Metrics:} In our study, mean absolute error (MAE) and root mean square error (RMSE) were used to measure the performance of experiment results. The definitions of the two metrics are: $\text{MAE}=\frac{1}{N}\sum^{N}_{i=1}|(y_i-\hat{y}_i)|$ and $\text{RMSE}=\frac{1}{N}\sum^{N}_{i=1}{(y_i-\hat{y}_i)}^2$, where $y_i$ and $\hat{y}_i$ are the real value and the predicted value, respectively.
% \end{itemize}

\subsection{Results and Analysis}
Table \ref{tab:X1V1} summarizes the results of all ST-GNN models based on the two datasets. The prediction horizon ranged from three time steps to 24 steps. The best evaluation results are highlighted in boldface. The number of highlighted values is also recorded (i.e., the winning counts) to compare the performance of different models.

\begin{figure}[t]
      \centering
            \begin{subfigure}{0.22\textwidth} % width of left subfigure
                  \includegraphics[width=\textwidth]{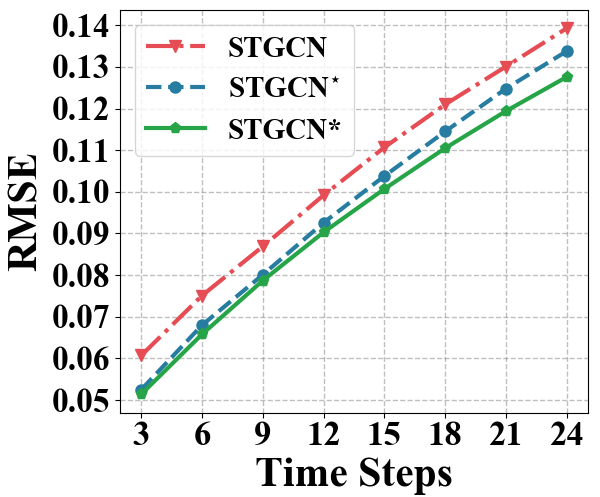}
                  \caption{STGCN} % subcaption
                  %\label{x7parking24}
            \end{subfigure}
            \begin{subfigure}{0.22\textwidth} % width of left 
                  \includegraphics[width=\textwidth]{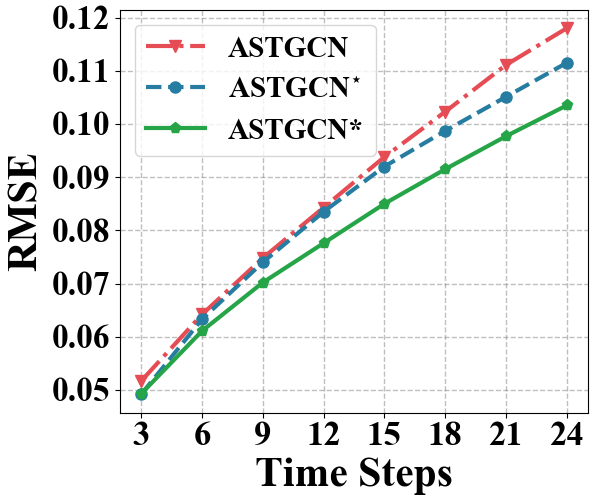}
                  \caption{ASTGCN}
                  %\label{x7air12}
            \end{subfigure}
            \begin{subfigure}{0.22\textwidth} % width of left subfigure
                  \includegraphics[width=\textwidth]{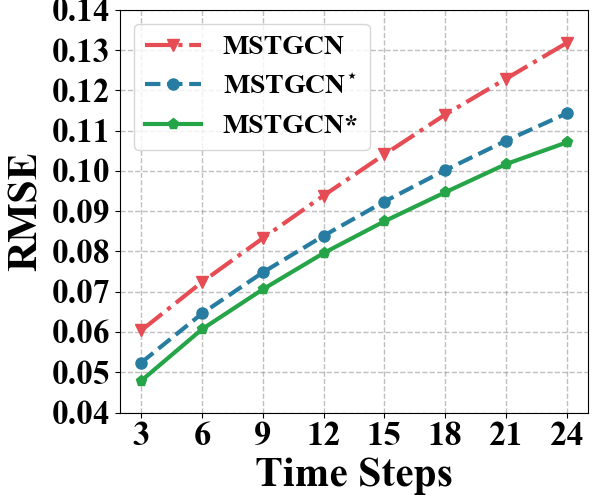}
                  \caption{MSTGCN} % subcaption
                  %\label{x7air24}
            \end{subfigure} 
            \begin{subfigure}{0.22\textwidth} % width of left subfigure
                  \includegraphics[width=\textwidth]{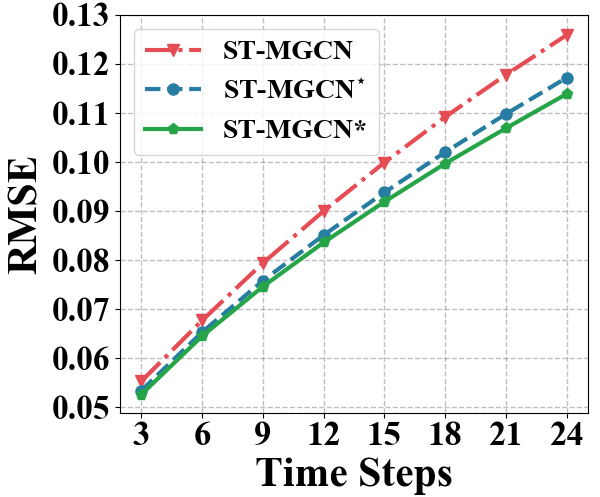}
                  \caption{ST-MGCN} % subcaption
                  %\label{x7air24}
            \end{subfigure} 
            \begin{subfigure}{0.22\textwidth} % width of left subfigure
                  \includegraphics[width=\textwidth]{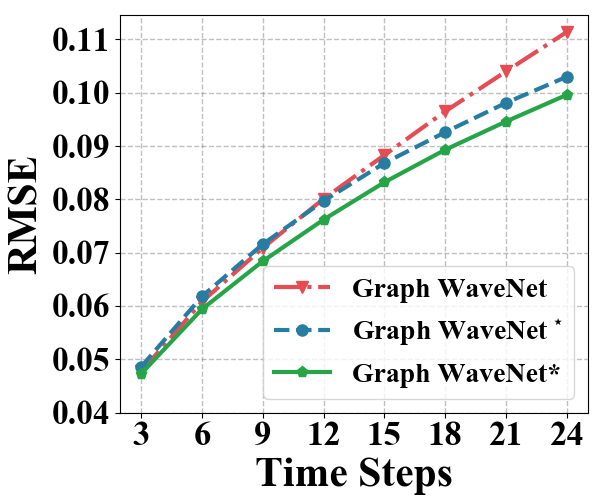}
                  \caption{Graph WaveNet} % subcaption
                  %\label{x7air24}
            \end{subfigure}
            \caption{The predicted RMSE of each model on the \textit{Parking} dataset over all time steps. The red line indicates the prediction errors of vanilla ST-GNN models, the blue line ($^{\star}$) shows the results of models using the proposed graph fusion methods but without SG-ATT, and the green line ($^*$) shows the results of models with multiple graphs with the proposed dynamic graph fusion approach.} % caption for whole figure
            \label{x2}
  \end{figure}

  \begin{figure}[!htbp]
      \centering
            \begin{subfigure}{0.43\textwidth} % width of left 
                  \includegraphics[width=\textwidth]{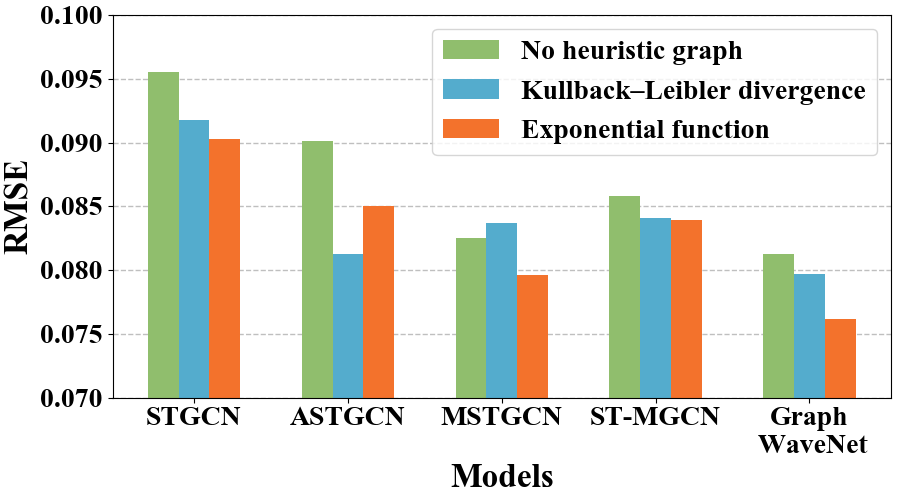}
                  \caption{Prediction horizon 12}
                  \label{x7parking12}
            \end{subfigure}
            \begin{subfigure}{0.43\textwidth} % width of left subfigure
                  \includegraphics[width=\textwidth]{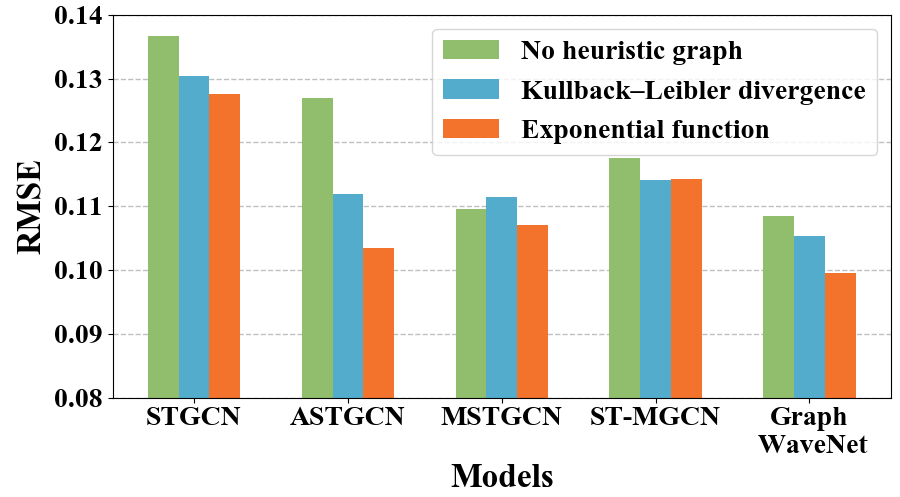}
                  \caption{Prediction horizon 24} % subcaption
                  \label{x7parking24}
            \end{subfigure}
            \caption{The Performance of models in the \textit{Parking} dataset. Each model is tested without heuristic graph or with heuristic graphs generated by the KL divergence or the exponential approximation function.} % caption for whole figure
            \label{x7}
\end{figure}

\begin{table}[t]
      \centering
      \scalebox{0.75}{
        \begin{tabular}{c|c|c|c|c|c|c|c}
        \toprule
        \multicolumn{3}{c|}{Model} & STGCN & ASTGCN & MSTGCN & ST-MGCN & \multicolumn{1}{p{4.19em}}{\makecell[c]{Graph\\WaveNet}} \\
      \specialrule{.08em}{.4ex}{.65ex}
        \multirow{4}[4]{*}{\begin{sideways}MAE\end{sideways}} & \multirow{2}[2]{*}{12} & $\dag$ & 0.0648  & 0.0648  & 0.0579  & 0.0612  & 0.0579  \\
              &       & $\ddag$ & \textbf{0.0640} & \textbf{0.0565} & \textbf{0.0575} & \textbf{0.0599}  & \textbf{0.0545} \\
    \cmidrule{2-8}          & \multirow{2}[2]{*}{24} & $\dag$ & 0.0961  & 0.0961  & 0.0809  & 0.0857  & 0.0818  \\
              &       & $\ddag$ & \textbf{0.0934} & \textbf{0.0768} & \textbf{0.0794} & \textbf{0.0836}  & \textbf{0.0724} \\
      \specialrule{.08em}{.4ex}{.65ex}
        \multirow{4}[4]{*}{\begin{sideways}RMSE\end{sideways}} & \multirow{2}[2]{*}{12} & $\dag$ & 0.0910  & 0.091  & 0.0882  & 0.0860  & 0.0818  \\
              &       & $\ddag$ & \textbf{0.0903} & \textbf{0.0776} & \textbf{0.0796} & \textbf{0.0836}  & \textbf{0.0762} \\
    \cmidrule{2-8}          & \multirow{2}[2]{*}{24} & $\dag$ & 0.1306  & 0.1178  & 0.1099  & 0.1281  & 0.1129  \\
              &       & $\ddag$ & \textbf{0.1276} & \textbf{0.1035} & \textbf{0.1071} & \textbf{0.1139}  & \textbf{0.0996} \\
        \bottomrule
        \end{tabular}%
      }
      \caption{The predicted RMSE of each model in the \textit{Parking} dataset. `$\dag$' and `$\ddag$' indicate the ST-GNN model that applies multi-graph fusion using the functionality graph proposed by \protect\cite{geng2019spatiotemporal} or the proposed functionality graph, respectively.}
      \label{tab:X4}%
\end{table}%

%\subsubsection{Long-Term Spatio-Temporal Forecasting}
In the first experiment, we aimed to provide an overall evaluation of the performance of the constructed graphs and the fusion approaches. We compared results between the existing ST-GNN without the proposed fused multi-graphs, and the results with the proposed multi-graph mechanism.  

Table \ref{tab:X1V1} shows the following: \textbf{(1)} When the proposed dynamic multi-graph fusion approach (marked with `*') was used, the prediction performances significantly improved. For example, when the STGCN method was used, our method had an average RMSE decrease of 9.5\% (over all prediction horizons).
% ; when using the ST-MGCN, our method had an average RMSE decrease by 7.7\% (over all prediction horizons). 
This indicates that our multi-graph fusion methods can extract more information and are effective for various ST-GNN models. \textbf{(2)} When the same ST-GNN methods are used, our proposed methods outperform the original ones in winning counts under all circumstances, which demonstrates the strong generality of our approach. \textbf{(3)} The results illustrate that our model is more suitable for the LSTF problem. Specifically, with the increase in prediction horizons, the gaps between vanilla ST-GNN models and our proposed models become larger. Figure \ref{x2} illustrates the trends of the proposed model and existing ST-GNN models with various prediction horizons. We found that the performance of the proposed models (green line) did not show a significant drop with the increasing prediction horizons while existing ST-GNN models (red line) underperformed in a long-run prediction.

% \subsection{Ablation Study: How well the proposed new graph and fusion approaches work?}
\subsection{Ablation Study}
To validate the performance of each component, we further conducted ablation studies on the \textit{Parking} dataset.

%\subsubsection{The Performance of Functionality Graphs}
% We conducted experiments with all ST-GNN models using the same settings but different functionality graphs with prediction horizons 12 and 24. 
% Although all the experiments used the same settings, how to construct the functionality graph differed. 
\paragraph{The Performance of Functionality Graphs.} Table \ref{tab:X4} shows that \textbf{(1)} most ST-GNN models using the proposed functionality graph (marked with `${\ddag}$') outperformed those using the functionality graph proposed by \cite{geng2019spatiotemporal}. \textbf{(2)} The results using the proposed functionality graph showed less drop when the prediction horizons changed from 12 to 24, which suggests that our proposed functionality graph performs well in LSTF tasks. 

%\subsubsection{The Performance of Heuristic Graph}
% We conducted experiments with all ST-GNN models using different heuristic graphs with prediction horizons 12 and 24, comprising graphs without heuristic graph, heuristic graphs generated by KL divergence and exponential approximation function. 
\paragraph{The Performance of Heuristic Graph.} Figure \ref{x7} shows that graphs generated by exponential approximation function in general outperformed other approaches with prediction horizons 12 and 24, while graphs generated by the KL divergence outperformed graphs without heuristic graphs. 
% \textbf{(2)} Although ST-GNN models with prediction horizons 12 performed better than prediction horizons 24, the performance differences on graphs generated by exponential approximation function were significantly less compared with graphs generated by KL divergence and without heuristic graph.  

%\subsubsection{The Performance of SG-ATT}
\paragraph{The Performance of SG-ATT.} Figure~\ref{x2} shows the performance of the framework with (marked with `$^*$') and without SG-ATT (marked with `$^{\star}$'). We observe that the SG-ATT mechanism contributes considerably to the proposed framework, especially in long-term prediction.
% \subsubsection{The Performance of Different Graph Fusion Approaches}
% In this study, we compared the performance of two different graph fusion approaches: the graph-wise fusion approach and the node-wise fusion approach. To validate the generality of these two graph fusion approaches, we tested them on four different ST-GNNs with two different prediction horizons 12 and 24 time steps. Figure \ref{x5} shows that the node-wise graph fusion approach outperformed the graph-wise one with all ST-GNN models, especially in predicting 24 time steps, which suggests that the node-wise graph fusion approach has potential in the LSTF problem. 

\section{Related Work}
Graph convolution networks (GCN) attracts much attention in spatio-temporal data prediction tasks recently. Bruna \textit{et al.} \cite{bruna2013spectral} proposed convolutional neural networks on graphs for the first time, which Defferrard \textit{et al.} \cite{defferrard2016convolutional} extended using fast localized convolutions. Using graph-based approaches, we can easily model spatial data. However, the observation from a single graph usually brings bias, while multiple graphs can offset and attenuate the bias. Chai \textit{et al.} \cite{chai2018bike} designed a multi-graph convolutional network for bike flow prediction. Geng \textit{et al.} \cite{geng2019spatiotemporal} encoded non-Euclidean pair-wise correlations among regions into multiple graphs and then modeled these correlations using multi-graph convolution for ride-hailing demand forecasting. Lv \textit{et al.} \cite{lv2020temporal} encoded non-Euclidean spatial and semantic correlations among roads into multiple graphs for traffic flow prediction. However, the relationships among graphs are ignored. Moreover, the input graphs are fixed and cannot be adapted to change during training and long-term temporal information is rarely considered.

\section{Conclusion}
In this paper, we try to solve the LSTF problem with multi-graph neural networks. We propose two new graphs to extract heuristic knowledge and contextual information from spatio-temporal data. Specifically, we designed a heuristic graph to capture the long-term pattern of the data and a functional similarity graph to represent the similarity of functionality between two areas. To align nodes in graphs and timestamps, we designed a dynamic graph multi-graph fusion module and fed them to various graph neural networks.
Extensive experiments on real-world data demonstrated the effectiveness of the proposed methods for enhancing the prediction capacity in LSTF problems. In the future, we will apply the proposed framework to additional graph-based applications.

\section*{Acknowledgments}
This work was made possible, in part, by grant NPRP No. NPRP12S-0313-190348 from the Qatar National Research Fund (a member of The Qatar Foundation). The statements made herein are solely the responsibility of the authors.
\bibliographystyle{named}
\bibliography{ijcai22_main}

\end{document}